\documentclass[preprint,review,12pt]{elsarticle}

\usepackage{amssymb}
\usepackage{lineno}

\usepackage[hang,small,bf]{caption}
\usepackage[subrefformat=parens]{subcaption}
\usepackage{array}
\usepackage{bm}
\usepackage{amsmath}

\usepackage{color}

\newcommand{\black}[1]{\textcolor{black}{#1}}

\journal{Pattern Recognition}

\begin{document}

\begin{frontmatter}

\title{Deep Image Compression Using Scene Text Quality Assessment}

\author[inst1]{Shohei Uchigasaki}
\ead{uchiga@iic.ecei.tohoku.ac.jp}

\author[inst1]{Tomo Miyazaki\corref{cor1}}
\ead{tomo@tohoku.ac.jp}
\cortext[cor1]{Corresponding author}

\author[inst1]{Shinichiro Omachi}
\ead{machi@ecei.tohoku.ac.jp}

\affiliation[inst1]{
organization={Graduate School of Engineering, Tohoku University},
addressline={6-6-05, Aoba Aramaki, Aoba}, 
city={Sendai},
postcode={980-8579}, 
state={Miyagi},
country={Japan}}

\begin{abstract}
Image compression is a fundamental technology for Internet communication engineering. However, a high compression rate with general methods may degrade images, resulting in unreadable texts. In this paper, we propose an image compression method for maintaining text quality. We developed a scene text image quality assessment model to assess text quality in compressed images. The assessment model iteratively searches for the best-compressed image holding high-quality text. Objective and subjective results showed that the proposed method was superior to existing methods. Furthermore, the proposed assessment model outperformed other deep-learning regression models.
\end{abstract}




\end{frontmatter}


\section{Introduction}
\begin{figure}[t] \centering
\begin{tabular}{cc}
\includegraphics[width=2.4in]{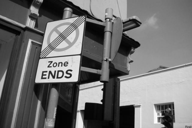} &\includegraphics[width=2.4in]{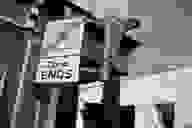}\\
(a) Original Image &(b) Image reconstructed using JPEG
\end{tabular}
\caption{Results of high-compression image compression.} \label{fig:1-1}
\end{figure}

The demand for higher image compression has increased due to the growth of communication traffic in recent years. Image compression is an essential technique for transmitting important content. One of the important contents in images is text, such as on signboards and book covers. However, traditional image compression methods such as JPEG~\cite{JPEG} may generate block noise when the compression ratio is high, resulting in unreadable text. Generally, JPEG images at high compression lost the readability of the text significantly, as shown in Figure~\ref{fig:1-1}.

An image compression technique that can retain text quality even under high compression is required. Nevertheless, image compression focusing on text quality has considerable room for development. A fundamental technique to evaluate a compressed image's quality is assessing text quality assessment. It would be inefficient for a human to evaluate the readability of all compressed images. By contrast, only a human can assess the readability of text. Two types of images contain text: scene images, such as landscapes and buildings, and document images, such as books and documents. Models for assessing text quality in scene images are yet to be studied.

In this study, we propose a model for assessing text quality in scene images using deep learning. Using only the image features obtained from convolution networks makes text readability assessment difficult. Therefore, we use a text recognition model to obtain informative features for the assessment. A text recognition model is similar to the human eye in text understanding. The proposed method assesses text quality using the image features and the text probabilities predicted by a text recognition model. The text quality assessment model allows the evaluation of the text quality (i.e., readability) of a scene image without human intervention. In addition, we propose an image compression method to ensure text quality. We use a deep neural network to develop the proposed image compression method. The proposed method controls text quality by changing the image compression parameters based on the text quality assessed from the image compression results.

The main contributions of this paper are the following.
\begin{itemize}
      \item We propose a quality assessment model for scene text images using image features and the text probabilities predicted by a text recognition model.
      \item We develop a deep image compression method that ensures text quality by applying the proposed assessment model.
      \item The proposed assessment and image compression models outperform traditional methods in objective measures.
\end{itemize}

The experimental results show that the proposed text quality assessment model outperforms existing deep learning feature extractors. Furthermore, the image compression method that maintains text quality outperforms traditional image compression methods.

\section{Related work}
The proposed method has two modules: text quality assessment and image compression. Thus, we describe related works in the aspect of the two modules. 

\subsection{Text quality assessment}
\black{Image quality assessment is a fundamental technique in image processing and computer vision. There are representative works with excellent performance. A referenceless quality metric was developed for synthesized images produced by a depth image-based rendering~\cite{7995055}. The literature~\cite{6963384} discovered that saliency preservation effectively enhances automatic contrast. Gu et al. analyzed the parameters of an autoregressive model to develop a no-reference sharpness metric~\cite{7115084}. Also, an assessment model uses image features of picture complexity, content statics, global brightness quality, and sharpness~\cite{7938348}.
}

Broadly, we can divide text images into two types: document images and scene text images. Various methods assess text quality in document images. For example, machine learning methods assess text quality from features extracted by filtering and edge detection~\cite{Tayo}. In addition, deep learning algorithms, such as a convolutional neural network (CNN), have been used for text quality assessment. The first deep learning-based method divides a document image into small patches and assesses them using a CNN~\cite{DIQA1}. Variant methods extract patches from different images~\cite{DIQA2} and various scales~\cite{DIQA3}. Jemn et al. enhance text images using a generative adversarial network to improve character recognition performance~\cite{KHAMEKHEMJEMNI2022108370}.

Text quality assessment models for scene images have not yet been studied. Super-resolution (SR) methods for scene text images relate to text quality assessment since SR improves text quality. TextSR~\cite{TextSR} uses a text recognition model as a loss function for SR. Likewise, TBSRN~\cite{TBSRN} uses a feature map extracted from the middle layer of a text recognition model. TextGestalt~\cite{TextGestalt} calculates loss values using text strokes. TPGSR~\cite{TPGSR} includes a text recognition model in its network. TATT~\cite{TATT} combines a text recognition model and transformer. Inspired by~\cite{TATT}, we use the transformer network to develop a quality assessment model for scene text images.

\subsection{Image compression}
Image compression methods include traditional methods, such as JPEG~\cite{JPEG}, JPEG2000~\cite{JPEG2000}, and deep learning methods. 

Firstly, we introduce conventional methods. JPEG and JPEG2000 are the most widespread image compression methods. Another method is sparse coding~\cite{Skretting1}. However, the conventional methods cannot use region-of-interest (RoI) compression and suffer from block noise.

Deep image compression methods can be divided into two types: variable-rate image compression~\cite{VariableRate1,VariableRate2} and ROI compression~\cite{ROICoding1,ROICoding2}. The RoI methods use a binary mask to maintain high quality for specified regions while ignoring the quality of other regions. Song et al.~\cite{QmapComp} develop a method enabling variable-rate and RoI compression. The method~\cite{QmapComp} uses a quality map instead of a binary mask. The quality map can compress images in various bit rates and allows fine values of quality in ROI. In this study, we employed the deep image compression method~\cite{QmapComp} to control text quality.

\section{Proposed method}
\begin{figure}[!t] \centering
\includegraphics[width=\linewidth]{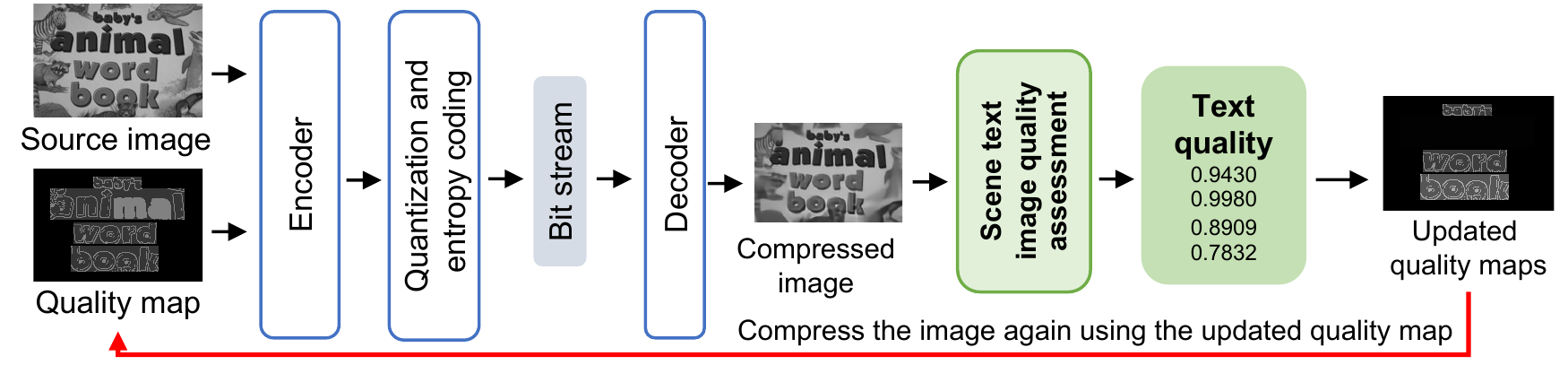}
\caption{Proposed framework} \label{3-1}
\end{figure}

We propose an image compression method for scene images. 
As shown in Figure~\ref{3-1}, the proposed method compresses images three times by updating the quality map. Then, we produce the final compressed image by selecting the compressed image that achieves the best text quality.
The main idea is to compress images by assessing text quality with the proposed scene text image quality assessment (STIQA) model. The proposed method combines variable-rate deep image compression~\cite{QmapComp} and the STIQA model. The proposed method defines the compression quality in each pixel. Thus, we can define a suitable quality map for scene text images. We update a quality map by assessing text quality using the STIQA model. This strategy provides a suitable text quality and compression for each text region.

\subsection{Scene text image quality assessment} 
\subsubsection{Overall framework of the STIQA model}
\begin{figure}[!t] \centering
\includegraphics[width=\linewidth]{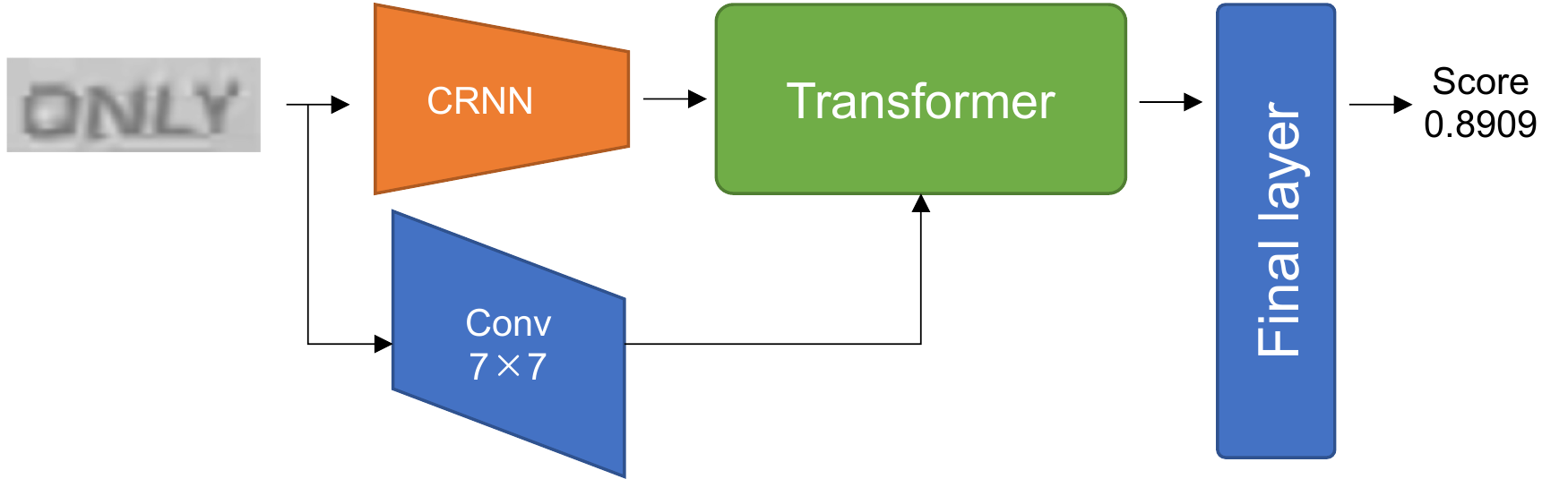}
\caption{Overall architecture of STIQA} \label{fig:3-2}
\end{figure}

An overall architecture of the STIQA model is shown in Figure~\ref{fig:3-2}. The STIQA model crops a text region from an image, resizing it to $32 \times 128$. We extract features from the text region using two networks. Firstly, we predict a sequence of probabilities for character categories using a convolutional recurrent neural network (CRNN)~\cite{CRNN}. The character categories include a–z, 0–9, and white space. Thus, the total number of character categories is 37. The second network extracts image features using a convolution layer with a window size of $7 \times7$. Then, we fuse the sequence and the image features using the transformer~\cite{Transformer} to produce features. The final layer comprises adaptive average pooling, fully connected layers, and a sigmoid activation function. The output is a predicted text quality ranging from [0, 1]. Higher value is high quality.

\subsubsection{Text image resizing}
The text regions are cropped from a scene image and are resized to $32 \times 128$. A standard rescale method, such as bilinear interpolation, distorts the aspect ratio of the original text. Therefore, we resize the text by expanding backgrounds if the text size is less than $32 \times 128$. We use the second-lowest pixel value among the four corner pixels, and Gaussian noise is then added to the pixel value. On the other hand, if the image size exceeds $32 \times 128$, we randomly crop the text to $32 \times 128$. As a result, the original text image can be input into the STIQA model.

\subsubsection{Transformer}
\begin{figure}[!t] \centering
\includegraphics[width=\linewidth]{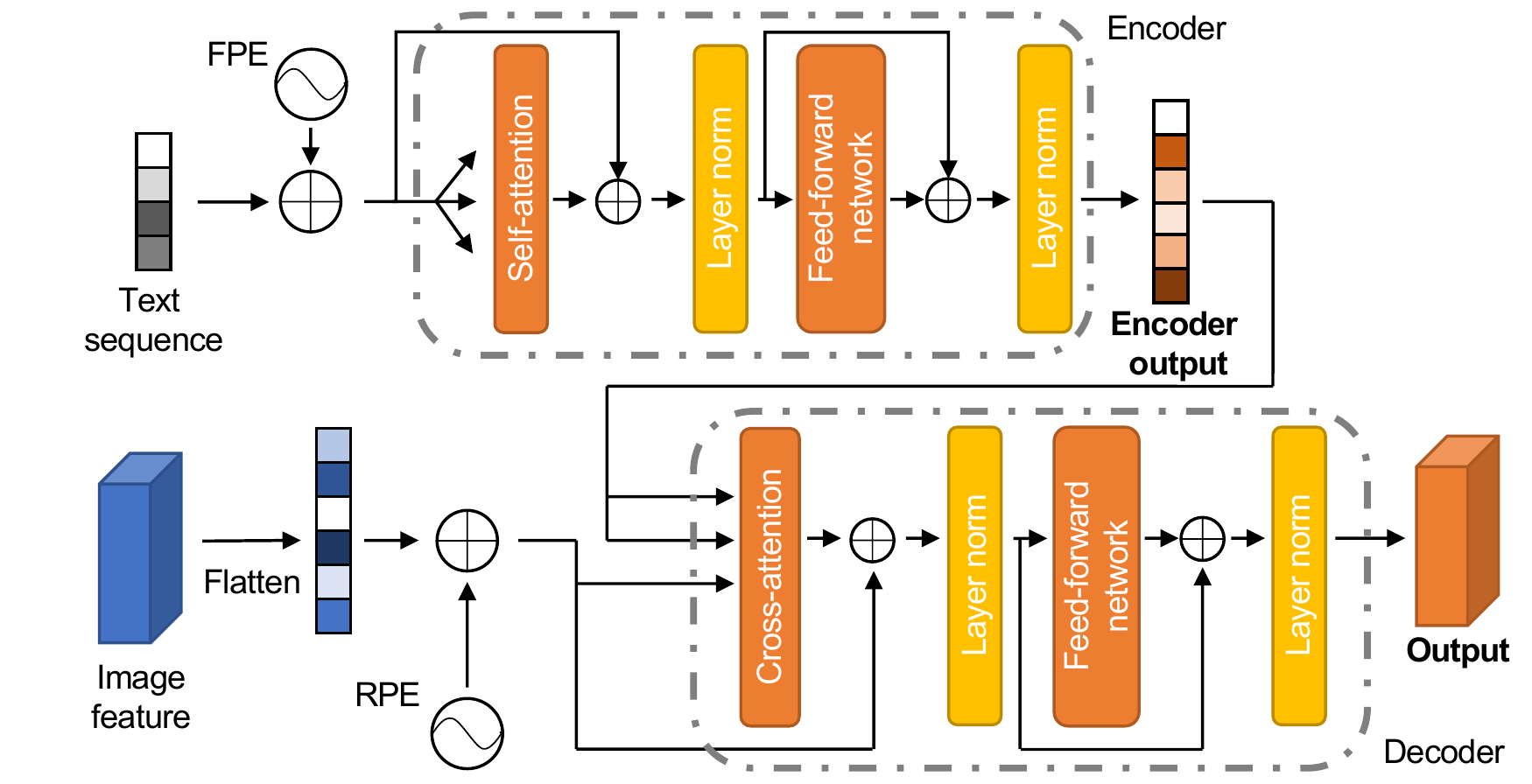}
\caption{Detailed architecture of the Transformer in the STIQA} \label{fig:3-3}
\end{figure}

We adopted the transformer~\cite{Transformer} to predict text quality. Figure~\ref{fig:3-3} illustrates the architecture. The Transformer can find the global relationships between two different features and output them as features. Convolution cannot deal with scene text images of various shapes and angles owing to its narrow effective range. Therefore, the Transformer with a global correlation is appropriate. The Transformer has an encoder and decoder. The positional encoding processes, called FPE and RPE, are carried out separately since the input does not consider the position or semantic order. Therefore, positional information must be added in advance.

The encoder extracts correlations and features from the text strings. In addition, the encoder reshapes the size of the features. The encoder embeds the sequence of character probabilities and the image features into the same embedding space. Specifically, the self-attention operation is used to extract correlations between the probabilities. The feed-forward network performs feature refinement.

The decoder transforms the image features into those that contain text information. Cross-attention is used to extract mutual features for the encoder output and image features. In other words, the decoder associates the text information output from the encoder with the corresponding positions in the image features. With the decoder, image features include text information, which strongly influences the text quality assessment.

\subsubsection{Loss function}
During training, the loss function uses the error between the ground truth and predicted labels. We use the L1 norm loss and the $\epsilon$-insensitive loss function. The $\epsilon$-insensitive loss function does not give a penalty if the error is within the allowed $\epsilon$ and gives a penalty if the error exceeds $\epsilon$. The formula for the $\epsilon$-insensitive loss function $L_\epsilon$ is defined in Eq.~\eqref{eq:loss:ep}. $GT$ represents the ground truth label and $Pred$ the predicted value. In this experiment, $\epsilon=0.1$. The reason for introducing the $\epsilon$-insensitive loss function is to penalize the learning process further if the estimation of text quality is largely incorrect. The overall loss function $L_{all}$ is defined by Eq.~(\ref{eq:loss:all}).

\begin{equation}
L_{\epsilon} = \left\{
\begin{array}{ll}
    0, &  |GT - Pred| < \epsilon \\
    |GT - Pred| - \epsilon,  & |GT - Pred| \geq \epsilon 
\end{array}
\right. \label{eq:loss:ep}
\end{equation}

\begin{equation}
    L_{all} = L_1 + L_{\epsilon} = |GT - Pred| + L_{\epsilon} \label{eq:loss:all}
\end{equation}

\subsection{Image compression methods considering text quality}
\subsubsection{Image compression using quality maps}
\begin{figure}[t] \centering
\includegraphics[width=\linewidth]{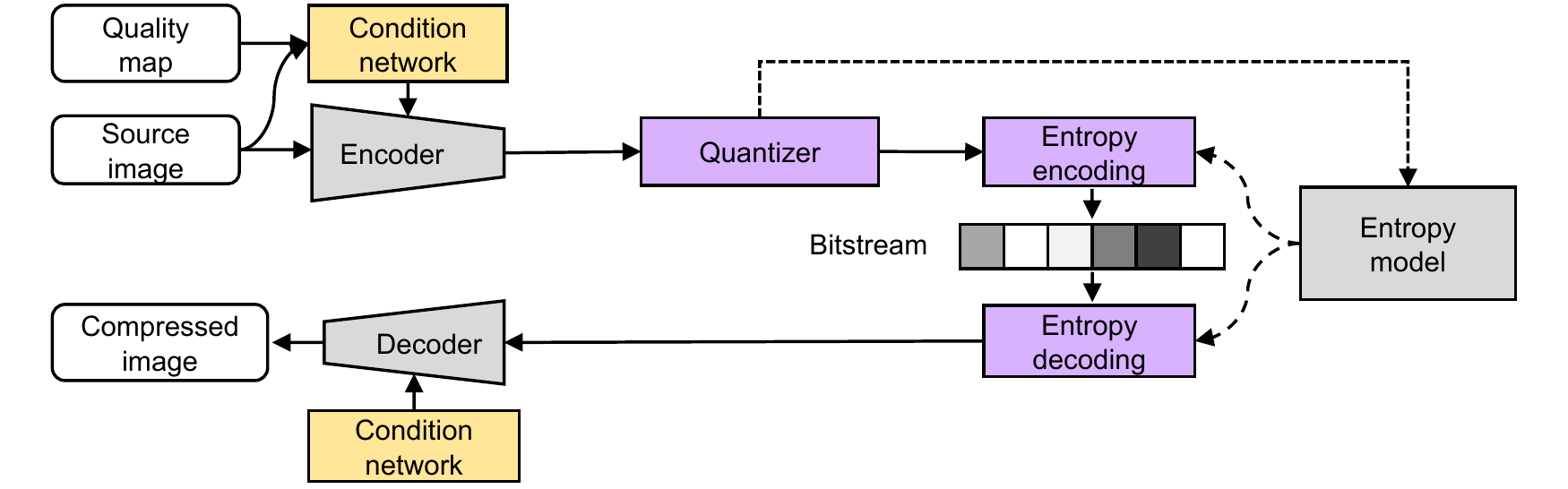}
\caption{The architecture of deep image compression model using quality maps} \label{3-4}
\end{figure}

The proposed image compression method uses variable-rate deep image compression~\cite{QmapComp}. The network is illustrated in Figure~\ref{3-4}. We can perform variable-rate image compression by using a pixel-by-pixel quality map. 

The SFT layer~\cite{SFT} is used in the condition network, encoder, and decoder. The others consist of convolution and normalization layers. The condition network fuses the information from a quality map using image features. Pixel-by-pixel quality maps enable high image quality in various areas, such as text and significant areas. Therefore, it can be used for a wide variety of tasks. The quality map has values ranging from [0, 1] to allow fine quality control. Henceforth, the value assigned to the quality map is denoted by the weight $k$.

A previous work~\cite{QmapComp} assigns the same weight value $k$ to all text regions in the image. However, the size and shape of the text vary significantly. Thus, the texts are different in readability. Therefore, we assigned different weights to each text region in this study.

\subsubsection{Quality map update}
\begin{figure}[t] \centering
\includegraphics[width=\linewidth]{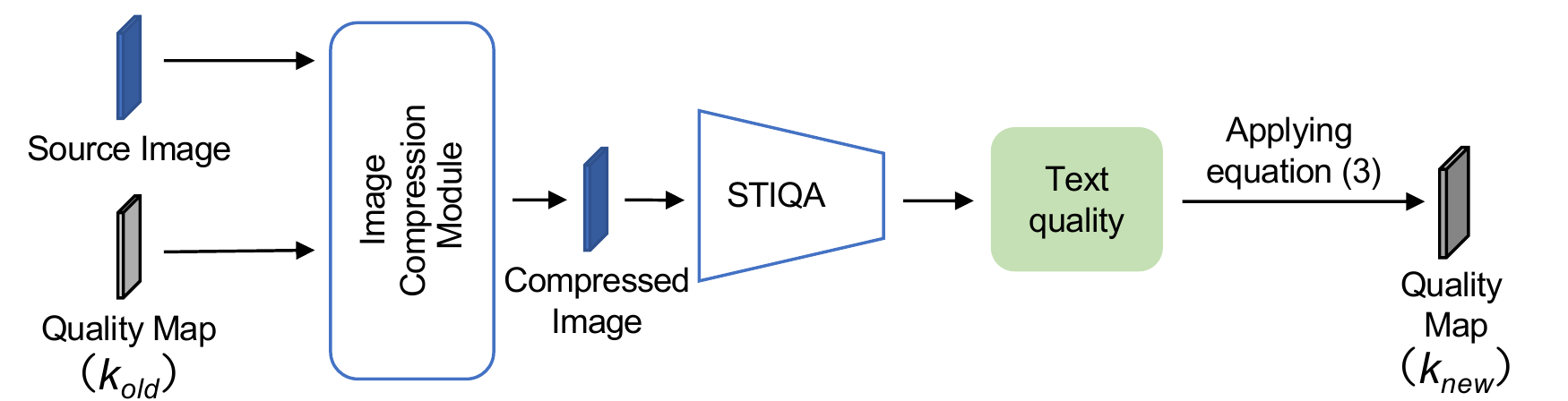}
\caption{Quality map update method} \label{fig:3-5}
\end{figure}

We propose an image compression method using a quality map to preserve text quality assessed by STIQA. The flow of the proposed image compression method is illustrated in Figure~\ref{fig:3-5}. This quality map is weighted differently for each text region.

The initial quality map has weights of 0.5 on pixels of character edges and inside. Otherwise, the weight is 0.2 for the background. High weights are assigned to the edges because character edges are the most critical factors for readability. We use the Canny method to extract edges.

We compress images and assess the quality of each text in the compressed image. We update weights $k_{new}$ by the distance between the assessment results and the target text quality. The update process is defined in Eq.~\eqref{eq:renew}. $k_{old}$ denotes the value of the previous weights, $\lambda$ is the hyperparameter that increases the update amount. $Score_{target}$ represnets the target text quality. $Score$ is the assessed text quality. We use  $\lambda=5$ and $Score_{target}=0.90$ in this work.

\begin{equation}
    k_{new} = k_{old} + \lambda (Score_{target} - Score)  \label{eq:renew}
\end{equation}

We iteratively perform image compression using the updated quality map. In this paper, the repetition is three times. Consequently, we can obtain a quality map that achieves high compression while maintaining text quality.

\section{Experiments}
We conducted experiments to evaluate the proposed STIQA and the image compression method. Firstly, we described the datasets and training details. Then we showed the experimental results of the proposed STIQA and the image-compression methods.

\subsection{Dataset}
The experiments used two datasets: ICDAR2013~\cite{ICDAR2013} and TextOCR~\cite{TextOCR}. The ICDAR2013 dataset is the focused scene text dataset of the Robust Reading Competition in ICDAR 2013, which is widely used in scene text analysis. TextOCR is a benchmark dataset for text recognition of arbitrarily shaped scene text. ICDAR2013 and TextOCR contain 229 and 21,778 training images, respectively.

We determined the quality of the text regions in 163 test images from the ICDAR2013 dataset by human annotation. In the human annotation task, 15 participants assessed the readability of the text in five grades. The answers were averaged and normalized to [0, 1]. We randomly selected 163 images from the test images of ICDAR2013 to alleviate annotation work.

We used a text recognition model, ABINet~\cite{ABINet}, to determine the quality of the text regions in the training images of ICDAR2013 and TextOCR since human annotation is infeasible for a large number of training images. We assumed that the human and the model are relevant if the model is accurate. Specifically, text quality $q$ is determined by Eq.~\eqref{eq:text:quality}. We used two measurements: the prediction confidence $c$ and accuracy $a$. ABINet predicted a set of probabilities $\bm{P} = ( \bm{p}_1, \cdots, \bm{p}_n )$ for $n = \mid \bm{p} \mid$ characters in a text region. $\bm{p}_i = (p_1,\cdots, p_{37})$ represents the output of a softmax function for the 37 classes. In addition, we measured the Levenshtein distance $d_\text{lev}$ between the text of the ground truth $t_g$ and the prediction $t_p$.
\begin{align}
q &= \frac{1}{2} ( c + a ), \label{eq:text:quality} \\
c &= \frac{1}{\mid \bm{P} \mid} \sum_{\bm{p} \in \bm{P}} \max(\bm{p}), \\
a &= 1 - \frac{d_\text{lev}( t_g, t_p )}{n} .
\end{align}

\subsection{Training details}
The STIQA model was trained using text regions cropped from highly compressed JPEG images of ICDAR2013 training dataset. There were 844 text regions in 229 scene images. The text regions were split into training and validation sets at an 80:20 ratio. The scene images were converted into grayscale and resized to $192 \times 128$ pixels. The purpose of resizing the images was to verify the performance of the quality assessment, even for small text regions. We used a batch size of 16 and 500 epochs, a learning rate of $10^{-3}$, and the Adam optimizer. 

The training images of the TextOCR dataset were used to train the proposed image compression method. Briefly, the same conditions as described in~\cite{QmapComp} were followed. The images of the TextOCR dataset were randomly cropped to $256 \times 256$ pixels.

\subsection{Experiments on image compression}
We compressed the test images and evaluated the quality of text regions. The proposed image compression method is compared with traditional image compression methods. We conducted a comparative experiment using objective and subjective evaluation metrics.

For comparison, we used the existing JPEG, JPEG2000, and a naive method using sparse coding~\cite{Skretting1}. We developed the naive method using optional weights in sparse coding. The naive method varied the weights and applied them to the text regions according to the text quality obtained by the STIQA. Therefore, the comparisons ensured fairness between the naive and the proposed methods. Furthermore, we used Qmap~\cite{QmapComp} to compress the images. All methods compressed the images at the same bit rate.

The evaluation metrics are PSNR, SSIM~\cite{SSIM}, and LPIPS~\cite{LPIPS}. In addition, the performance of a text detection model, TextFuseNet~\cite{TextFuseNet}, was used to evaluate the compressed scene text images. We use precision, recall, and f-measure. The evaluation metrics were calculated for compressed images at almost identical bit rates.

\subsubsection{Main results on image compression}
The image compression results were presented in Table~\ref{tab:ImgCodComp}. The results were averaged over the test images in ICDAR2013. The results showed that the proposed method was superior for all the metrics. The proposed method obtained the second-best results for the entire image. The proposed image compression method outperformed traditional image compression methods, such as JPEG. Furthermore, the proposed method is the best in terms of text quality and text detection metrics. Therefore, the results verified the effectiveness of the proposed image compression method.

\begin{table}[t] \centering
\caption{Results on image compression (Bold and underline represent the best and the second best, respectively.)} \label{tab:ImgCodComp}
\begin{tabular}{l|ccc|ccc|ccc}
&\multicolumn{3}{c|}{Entire image quality} &\multicolumn{3}{c|}{Text quality} &\multicolumn{3}{c}{Text detection}\\
         &PSNR &SSIM &LPIPS &PSNR &SSIM &\black{LPIPS} &P    &R    &F \\ \hline
JPEG     &23.4 &0.65 &0.29  &19.6 &0.56 &\black{0.27}  &0.28 &0.19 &0.22 \\
JPEG2000 &26.9 &0.77 &0.25  &22.5 &0.72 &\black{0.18}  &0.60 &0.51 &0.54 \\
Naive
         &25.6 &0.71 &0.23  &\underline{27.0} &\underline{0.89} &\black{\underline{0.07}} &\underline{0.75} &\underline{0.65} &\underline{0.68} \\
Qmap~\cite{QmapComp} 
         &\textbf{31.2} &\textbf{0.88} &\textbf{0.13}  &26.2 &0.85 &\black{0.11} &0.68 &0.61 &0.63 \\
Proposed &\underline{29.7} &\underline{0.85} &\underline{0.17} &\textbf{28.7} &\textbf{0.93} &\black{\textbf{0.05}} &\textbf{0.79} &\textbf{0.74} &\textbf{0.76}
\end{tabular}
\end{table}

\begin{figure}[t] \centering
\begin{tabular}{ccc}
\includegraphics[width=1.7in]{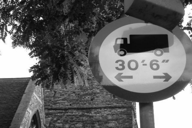} &\includegraphics[width=1.7in]{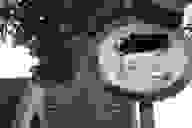} &\includegraphics[width=1.7in]{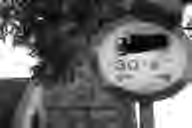} \\
(a)Original &(b) JPEG &(c) JPEG2000 \\
\includegraphics[width=1.7in]{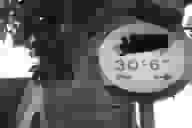} 
&\includegraphics[width=1.7in]{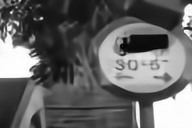}
&\includegraphics[width=1.7in]{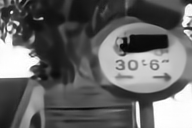} \\
(d) Naive &(e) Qmap &(d) Proposed
\end{tabular}
\caption{Compressed images} \label{fig:4-1}
\end{figure}

\black{Qmap outperformed ours in overall image quality because Qmap was optimized for overall images, whereas our method focuses on text readability. Image compression is to determine a bitstream for a set of pixel values, as shown in Figure~\ref{3-4}. Qmap determines a bitstream using uniform weights for all pixels, i.e., the overall image. Thus, the overall quality is favorable. However, text quality greatly deteriorates since the number of text pixels is smaller than the other pixels. In contrast, our method modifies weights by assigning more bits to text, resulting in a bitstream maintaining text readability. As shown in Figure~\ref{fig:4-1}, although overall image quality is reduced slightly, text quality improved greatly.} 

Some of the compressed images are shown in Figure~\ref{fig:4-1}. JPEG and JPEG2000 were significantly degraded in both background and text quality. In addition, the naive method degrades only the background. By contrast, the proposed method achieves a higher image quality for text regions and a higher image quality for background regions compared to the other methods.

\subsubsection{Subjective evaluation results}
Subjective evaluation experiments were conducted using the proposed and naive methods. Fifteen participants chose better readability between two compressed images in terms of the entire image and the text region. Each participant had 50 pairs.

The results are presented in Table~\ref{tab:Shukan}. The proposed method is superior for the entire image and text regions. The proposed method obtained more than 66\% of the votes for the text region. These results demonstrate the capability of this method in maintaining text readability.

\begin{table}[t] \centering
\caption{The ratios of Subjective evaluation on comparison of compressed images by the Naive and the proposed methods} \label{tab:Shukan}
\begin{tabular}{l|r|r} 
Answer &Entire image & Text region \\ \hline
Same quality &0.3(\%) & 28.4 (\%) \\
Naive is better &2.4 (\%) & 5.3 (\%) \\
Proposed is better & \textbf{97.3} (\%) & \textbf{66.3} (\%)
\end{tabular}
\end{table}

\subsubsection{Hyper Parameter determination in image compression}
The effect of varying the hyperparameter $\lambda$ and the number of updates was also investigated. $\lambda$ is defined in Eq.~(\ref{eq:renew}) to determine the amount of quality map update. Table~\ref{tab:QmapParam} shows the bit rates (bpp) and the average score of the text qualities assessed using STIQA. The bit rates and STIQA scores increased according to $\lambda$ and the number of updates. The best STIQA score was obtained when $\lambda=5$. By considering the computing time, it was determined that three update cycles were sufficient.

\begin{table}[t] \centering
\caption{Bit rates and STIQA scores at various hyperparameters} \label{tab:QmapParam}
\begin{tabular}{r|cc|cc}
 &\multicolumn{2}{c|}{3 updates} &\multicolumn{2}{c}{10 updates} \\
$\lambda$       &bpp  &STIQA &bpp  &STIQA \\ \hline
$10^{-1}$       &0.18 &0.57  &0.18 &0.59 \\
$10^0$          &0.19 &0.59  &0.22 &\textbf{0.62} \\
$5 \times 10^0$ &0.22 &\textbf{0.62}  &0.22 &\textbf{0.62} \\
$10^1$          &0.23 &\textbf{0.62}  &0.23 &\textbf{0.62} \\
$10^2$          &0.23 &0.61  &0.23 &0.61
\end{tabular}
\end{table}

\subsection{Experiments on text quality assessment}
We evaluated the STIQA model and other methods. The comparison methods used in this experiment were the ResNet-34~\cite{ResNet}, VGG-13~\cite{VGG}, EfficientNet-B3~\cite{EfficientNet}, and Document Image Quality Assessment (DIQA)~\cite{DIQA2}. We trained models to regress the text quality assessment.

\subsubsection{Text quality assessment results}
The results are presented in Table\ref{tab:CompQA}. Three evaluation metrics were used: mean absolute error, Spearman’s rank correlation coefficient, and Pearson’s correlation coefficient. \black{We used the indicators to show a relationship between humans and our model in text quality assessment. Specifically, MAE shows the gap between a model and humans. Pearson’s correlation coefficient uses scores of text quality to evaluate a linear relationship. Spearman’s rank correlation coefficient uses ranks among data to verify a monotonic relationship. Thus, we can comprehensively evaluate text quality assessment models using the three indicators.}

The results showed that STIQA performed the best on the test dataset with text quality labels obtained from subjective evaluation experiments. The main difference between STIQA and other methods is the existence of a text recognition model, which may play a significant role in text quality assessment.

\begin{table}[t] \centering
\caption{Results on text quality assessment} \label{tab:CompQA}
\begin{tabular}{r|ccc} 
Model &MAE &Spearman &Pearson \\ \hline
EfficientNet~\cite{EfficientNet} &0.31 &0.27 &0.24 \\
VGG-13~\cite{VGG} &0.30 &0.25 &0.28 \\
Res-34~\cite{ResNet} &0.28 &0.37 &0.34 \\
DIQA~\cite{DIQA2} &0.22 &0.43 &0.47 \\
STIQA (Proposed) &\textbf{0.20} &\textbf{0.55} &\textbf{0.59} 
\end{tabular}
\end{table}

\subsection{\black{Ablation study on STIQA}}
\black{We conducted an ablation study on the proposed Scene Text Image Quality Assessment model (STIQA). The STIQA consists of three modules: CRNN, Conv layer, and Transformer as shown in Figure~\ref{fig:3-2}. The CRNN extracts text features by predicting probabilities for character categories. Also, the Conv layer extracts image features. The Transformer fuses the text and image features to produce features. The final layer is a fully connected layer to predict text quality.}
\black{The results are shown in Table~\ref{tab:STIQA:ablation}. The CRNN model obtained relatively good performance. In contrast, the other models were superior. The full STIQA achieved the best performance. Comparing CRNN + Tans and the full STIQA, the Conv layer significantly improved performance. Therefore, the image features are essential to the proposed text quality assessment model.}

\begin{table}[t] \centering
\caption{\black{Ablation results of STIQA in text quality assessment}} \label{tab:STIQA:ablation}
\begin{tabular}{l|ccc} 
Model &MAE &Spearman &Pearson \\ \hline
CRNN &0.218 &0.40 &0.46 \\
CRNN + Transformer &0.203 &0.47 &0.55 \\
CRNN + Transformer + Conv ( Full STIQA) &\textbf{0.197} &\textbf{0.55} &\textbf{0.59} 
\end{tabular}
\end{table}

\subsubsection{Parameter combination results}
We evaluated the impact of the combinations of the loss functions: the L1 loss and the $\epsilon$-insensitive loss. The results are presented in Table~\ref{tab:res:param:search}. The $L_\epsilon$ loss was better than the $L_1$ loss. In addition, the best results were achieved by using the two loss functions simultaneously, and the combination with an $\epsilon$ of 0.10 was the best overall.

\begin{table} \centering
\caption{Results on parameter combination in the proposed STIQA} \label{tab:res:param:search}
\begin{tabular}{ccl|ccc}
$L_1$ &$L_\epsilon$ &$\epsilon$ &MAE &Spearman &Pearson \\ \hline
$\surd$ & -       &NA   &0.212 &0.50 &0.57 \\
-       &$\surd$ &0.10 &0.210 &0.44 &0.50 \\
$\surd$ &$\surd$ &0.20 &0.214 &0.52 &0.55 \\
$\surd$ &$\surd$ &0.15 &0.206 &0.48 &0.56 \\
$\surd$ &$\surd$ &0.10 &\textbf{0.197} &\textbf{0.55} &\textbf{0.59}
\end{tabular}
\end{table}

\section{Conclusions}
We proposed a method for text quality-preserving image compression using a text quality assessment model. The proposed image compression method can produce readable scene text images by assessing the quality of the text regions in the images. Text quality was assessed using a transformer to consider global and local features extracted from the entire image and the text regions. The experimental results showed that the proposed method is superior to the JPEG and naive methods using sparse coding compression. In addition, the proposed text quality assessment model outperformed other regression models. Both objective and subjective evaluations showed that the proposed method is capable of text quality preservation during image compression.

\black{A promising application of the proposed method is video compression. We can efficiently preserve text information in a video conference, IoT, and sensor network. Besides, the proposed method can be applied to more applications if we extend the text quality assessment model to general objects.}

\black{Future work is to design an end-to-end model. The proposed model has two modules for image compression and text quality assessment. We trained the modules individually. Thus, we can improve the performance of the proposed method through end-to-end learning. The challenge is how to connect the two modules. A solution is text quality assessment using features encoded by the image compression model. This solution can facilitate learning the model and is beneficial to computational costs. 
}

\black{Constructing a practical system in combination with the text detection method is also essential for future work. Since it is difficult to evaluate the effectiveness of the proposed method when text detection errors occur, we conducted the experiments assuming that the text regions are known.
The proposed STIQA can evaluate text readability. Thus, we believe it is possible to improve the accuracy of text detection by evaluating the readability of the regions detected by text detection methods.}

\section*{Acknowledgments}
This work was partially supported by JSPS KAKENHI under Grants JP20H04201, JP22K12729, and JP22H00540.

 \bibliographystyle{elsarticle-num} 
 \bibliography{reference}

\end{document}